ORIGINAL RESEARCH

Open Access

# Comparison of machine learning methods for classifying mediastinal lymph node metastasis of non-small cell lung cancer from $^{18}$F-FDG PET/CT images

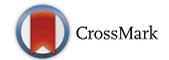

Hongkai Wang[1], Zongwei Zhou[2], Yingci Li[3], Zhonghua Chen[1], Peiou Lu[3], Wenzhi Wang[3], Wanyu Liu[4] and Lijuan Yu[3*]

## Abstract

**Background:** This study aimed to compare one state-of-the-art deep learning method and four classical machine learning methods for classifying mediastinal lymph node metastasis of non-small cell lung cancer (NSCLC) from $^{18}$F-FDG PET/CT images. Another objective was to compare the discriminative power of the recently popular PET/CT texture features with the widely used diagnostic features such as tumor size, CT value, SUV, image contrast, and intensity standard deviation. The four classical machine learning methods included random forests, support vector machines, adaptive boosting, and artificial neural network. The deep learning method was the convolutional neural networks (CNN). The five methods were evaluated using 1397 lymph nodes collected from PET/CT images of 168 patients, with corresponding pathology analysis results as gold standard. The comparison was conducted using 10 times 10-fold cross-validation based on the criterion of sensitivity, specificity, accuracy (ACC), and area under the ROC curve (AUC). For each classical method, different input features were compared to select the optimal feature set. Based on the optimal feature set, the classical methods were compared with CNN, as well as with human doctors from our institute.

**Results:** For the classical methods, the diagnostic features resulted in 81~85% ACC and 0.87~0.92 AUC, which were significantly higher than the results of texture features. CNN's sensitivity, specificity, ACC, and AUC were 84, 88, 86, and 0.91, respectively. There was no significant difference between the results of CNN and the best classical method. The sensitivity, specificity, and ACC of human doctors were 73, 90, and 82, respectively. All the five machine learning methods had higher sensitivities but lower specificities than human doctors.

**Conclusions:** The present study shows that the performance of CNN is not significantly different from the best classical methods and human doctors for classifying mediastinal lymph node metastasis of NSCLC from PET/CT images. Because CNN does not need tumor segmentation or feature calculation, it is more convenient and more objective than the classical methods. However, CNN does not make use of the import diagnostic features, which have been proved more discriminative than the texture features for classifying small-sized lymph nodes. Therefore, incorporating the diagnostic features into CNN is a promising direction for future research.

**Keywords:** Computer-aided diagnosis, Non-small cell lung cancer, Positron-emission tomography, Machine learning, Deep learning

* Correspondence: yulijuan2003@126.com
[3]Center of PET/CT, The Affiliated Tumor Hospital of Harbin Medical University, 150 Haping Road, Nangang District, Harbin, Heilongjiang Province 150081, China
Full list of author information is available at the end of the article

## Background

In recent years, diagnosis of non-small cell lung cancer (NSCLC) from PET/CT images became a popular research topic [1–4]. Many studies focused on assessing the efficacy of $^{18}$F-FDG PET/CT for diagnosing mediastinal lymph node metastasis [5–12], whereas the judgment of metastasis was mostly based on thresholding the image features such as maximum short diameter, maximum standardized uptake value ($SUV_{max}$), or mean standardized uptake value ($SUV_{mean}$). Due to the limited number of image features and the simplicity of feature thresholding strategy, the diagnostic power of PET/CT might not have been fully explored. According to a recent summary [13] of the past 10 years for mediastinal lymph node NSCLC diagnosis using $^{18}$F-FDG PET/CT, the median sensitivity was only 62%, which means a large portion of metastasis were false-negatively judged. To improve the diagnosis sensitivity of mediastinal lymph node NSCLC, more sophisticated classification strategy is needed and computerized machine learning algorithms could be of help.

Computer-aided diagnosis (CAD) methods of medical images have been developed for decades [14], but many methods were designed for imaging modalities other than nuclear medicine, such as X-ray, CT, MR, and ultrasound. It was not until the recent 5 years that PET/CT texture features attracted increased research attention for tumor diagnosis [15–18], radiotherapy response characterization [19], and treatment outcome prediction [20]. Texture features reflect the heterogeneity of tumor uptake which could be helpful for differential diagnosis. However, due to the influence of various factors including imaging protocol, lesion size, and image processing, the effectiveness of PET/CT texture features is still under argumentation [21], and standardization of texture feature calculation is highly required [22]. For mediastinal lymph node NSCLC, further study is needed to evaluate the diagnostic ability of PET/CT texture features.

Along with the development of computer hardware and the growth of medical image data, the applications of deep learning technique for medical image CAD became a hot research direction. This technique uses deep artificial neural networks to learn the image appearance patterns of interested objects based on a large training data set. Deep learning has been reported to significantly outperform classical machine learning methods for object detection and classification and has been increasingly used for medical image analysis [23]. So far, the applications of deep learning for medical images include the detection and segmentation of lesions from CT images [24–26], colonoscopy videos [27], and histopathology images [28, 29], but the applications on tumor diagnosis are limited [30]. Compared to classical machine learning methods, deep learning method does not require segmentation of tumor, it simplifies the analysis procedure and avoids subjective user bias. To the extent of our knowledge, there has not been any study using deep learning technique for the classification of mediastinal lymph node NSCLC from FDG PET/CT images.

Based on the above considerations, this study aimed to compare the performance of multiple machine learning methods for classifying mediastinal lymph node NSCLC from PET/CT images. The evaluated methods included both classical feature-based methods and the state-of-the-art deep learning approach. For the classical methods, the texture features were compared with the features used by human doctors for clinical diagnosis, such as tumor size, CT value, SUV, image contrast, and intensity standard deviation. The machine learning methods were also compared with human doctors, so as to evaluate the value of computerized methods for classifying mediastinal lymph node NSCLC from FDG PET/CT images.

## Methods

### Data resources

The study was approved by our institutional research ethics board. This was a retrospective per-node study. $^{18}$F-FDG PET/CT images of 168 patients were retrieved from our hospital database within the period from June 2009 to September 2014. Lobectomy combined with systematic hilar and mediastinal lymph node dissection was performed in our institute. The locations of removed lymph nodes were tracked on a per-station basis. We removed the lymph nodes located on groups 1, 2R, 3, 4R, 7, 8, and 9R in the right lung and groups 4L, 5, 6, 7, 8, and 9L in the left lung, and put the lymph nodes from the same nodal station into one sampling bag. Pathological diagnosis of each specimen bag was made by an oncological pathologist with 18 years of experience based on hematoxylin-eosin staining. From the 168 patients, 1397 lymph nodes were confirmed cancerous by pathology, and the number of negative and positive samples were 1270 and 127, respectively. Detailed information is listed in Table 1.

The PET/CT scans were applied within 1 week before surgery. All patients fasted for more than 4 h before the scan to keep the blood glucose below 6.0 mmol/L. The patients were intravenously injected with 300–400 MBq $^{18}$F-FDG of ≥97% radiochemical purity synthesized by the GE Minitracer cyclotron and Tracer Lab FX-FDG synthesizer. After 1-h resting following the injection, patients were scanned by the GE Discovery ST PET-CT scanner. Whole-body CT scan was performed under shallow breathing status. The CT scanner settings were 120 kV, 140 mA, 0.5 s per rotation, 1.25:1 pitch, 3.75-mm slice thickness, 1.37-mm in-plane spatial resolution, and 20–30-s scan time. PET scan was performed in a 3D acquisition mode also under shallow breathing status. Six or seven bed positions were scanned for each patient with 2.5-min emission time per bed position. PET images were reconstructed by iterative algorithm, using CT image for attenuation correction.

**Table 1** Patient and lymph node characteristic

| | |
|---|---|
| Patients number (male/female/total) | 91/77/168 |
| Patient ages (min/max/median) | 38/81/61 |
| Lymph nodes number (benign/malignant/total) | 1270/127/1397 |
| Lymph nodes short axis diameter (≤2/≤4/≤7/≤10/>10 mm) | 306/816/246/23/6 |

Based on the PET/CT images, diagnosis of lymph node metastasis status (positive or negative) was made by four doctors from our institute, two of whom with over 10 years experience. Final consensus was reached after the discussion of all doctors. The doctors made their diagnosis based on several factors including maximum short diameter, maximum standardized uptake value ($SUV_{max}$), mean standardized uptake value ($SUV_{mean}$), visual contrast between the tumor, and its surrounding tissues in the CT image, as well as the location in the lymph node map [31].

#### Machine learning methods

This study compared four mainstream classical machine learning methods and one deep learning method. The classical methods included random forest (RF), support vector machines (SVM), adaptive boosting (AdaBoost), and back-propagation artificial neural network (BP-ANN). We refer the readers to [32] for detailed introduction of the classical methods. The four classical methods were implemented using the functions of MATLAB R2014b. This study used 10 times 10-fold cross-validation to evaluate the machine learning methods. For each of the cross-validations, the optimal parameters of each method were determined based on the nine folds of training samples, via grid search in the parameter space. Therefore, each cross-validation might have slightly different optimal parameters, and the average optimal values are reported here. The random forest contained 100 decision trees on average. The depth of each tree was controlled by a minimum leaf size of 1, the number of features used for each decision split was set to the square root of the total feature number. SVM used a Gaussian radial basis function as the kernel function, and the sequential minimal optimization method to find the separating hyperplane, its average kernel size was 2.0. AdaBoost used shallow decision trees as the weak classifiers, it included 300 shallow decision trees with a maximum split number equal to 1, the average learning rate was 0.1. ANN used two hidden layers with 50 and 26 neurons for the first and second layer, respectively, there were 1000 epochs, and the average learning rate was 0.04.

The deep learning method was the convolutional neural network [33], which is a deep neural network dedicated for image classification, it is also named the ConvNets in some literatures. CNN has been proved to significantly outperform the classical machine learning methods for natural image classification. Unlike the classical methods which take the feature vectors as input, CNN takes an image patch of $n \times n$ pixels as input. CNN performs classification according to the appearance of the image patch; it learns the patterns of patch appearance from a large amount of training patches. The outputs of CNN are the scores for different classes, and the class with the highest score is deemed as the classification result. For our application, the input of CNN is a patch around the lymph node, the outputs are two scores of being benign and malignant.

The architecture of CNN mimics the structure of animal visual cortex. The input image patch is firstly passed to several consecutive layers that convolute and downsample the patch, followed by a flattening layer which stretches the patch into a feature vector. After the flattening layer, the subsequent layers (namely the fully connected layers and the softmax layer) convert the feature vector into the output scores. In recent years, several improved CNN architectures have been proposed, but their overall architecture kept similar. This study used the well-known AlexNet [34] architecture implemented using the Keras library for Python. To avoid overfitting to our data, the number of AlexNet layers was reduced to five. Our CNN also incorporated L2 normalization, ReLU activation function, dropout regularization, categorical cross entropy loss function, and Adadelta learning method. The choice of CNN architecture will be further explained in the "Discussion" section.

For both training and testing stages, the inputs of CNN were six axial image patches cropped from the CT and PET SUV images. The six patches included three patches for each image modality. The patches were cropped around the lymph node center and resampled into $51 \times 51$ pixels of 1.0-mm size. The three patches of each modality included one slice centered at the lymph node center and two others located 4 mm above and below the lymph node center. Our parallel patch configuration was different from [25] which used orthogonal patches of axial, coronal, and sagittal directions, because we found that parallel patches resulted in higher AUCs than the orthogonal patches.

To generate the patches, the center of each lymph node was specified by the doctor. This was the only step requiring user input. To cope with the subjective variance of the user input, as well as to expand the size of the training set, data augmentation was performed for the training data. The image patches were translated and rotated in 3D space around the lymph node center to generate more pseudo training patches. Each patch was translated along $x$, $y$, and $z$ axes by $N_t$ steps and rotated about the three axes by $N_r$ angles. In this study, the translation steps were [−2,0,2] pixels and the rotation angles were [−20°,0°,20°], i.e., $N_t = N_r = 3$. As a result, each sample was extended to $N_t \times N_t \times N_t \times N_r \times N_r \times N_r = 3^6 = 729$ samples after data augmentation; hence, the total sample size was $1397 \times 729 = 1{,}018{,}413$. Such data augmentation strategy was commonly used by the deep learning methods.

To train the network, the momentum update method [35] was used, with batch size of 64 and momentum coefficient of 0.9. The initial learning rate $\eta$ was 1e−9, with a decreasing rate $\gamma = 0.9$ for each iteration. After 10 epochs, $\eta$ decreased to a very small value close to 0, and the learning process converged to a local minima, which might not be the global minima. To step out of the local minima, we reactivated the learning process by setting $\eta$ back to a larger value ($\eta$ = 1e−11), then continue the learning with $\gamma = 0.95$ until convergence. We found such reactivation scheme essential for obtaining good CNN performance.

### Input features for classical methods

In this study, the input features for the classical methods included 13 diagnostic features and 82 texture features. Please note that these features are only used for the classical methods, because CNN does not take the features as input.

Definitions of the features are listed in Table 2. The term "diagnostic feature" is to represent the features used by human doctors for clinical diagnosis, such as tumor size, SUV, CT values, and image contrast. In Table 2, the features of $D_{short}$, area, and volume are related to lymph node size, $CT_{mean}$ corresponds to tissue density, and $SUV_{mean}$ and $SUV_{max}$ represent lymph node metabolism level. $CT_{contrast}$ is used to measure the density difference between the lymph node and its vicinity, since metastatic tumor tends to merge with its surrounding tissue. $SUV_{std}$ measures the variation of metabolism level inside the lymph node, because some malignant tumors may have necrotic cores. For the features of $CT_{mean}$, $CT_{contrast}$, $SUV_{mean}$, $SUV_{max}$, and $SUV_{std}$, both 2D and 3D versions were calculated. 2D versions were calculated based on the axial slice passing the lymph node center, and 3D versions were computed based on the reconstructed volumes. We incorporated the 3D features to compensate for the limitation of the doctors' visual inspection of planar images.

To calculate the 95 features, manual segmentation of the lymph nodes was performed by two doctors for all axial slices covering the entire node. The lymph nodes were delineated based on the fusion of PET and CT images because the two modalities compensated each other

Table 2 The image features used in this study. For the column of "image modality", the term "PET/CT" means the feature is calculated for both PET and CT

| Feature | Image modality | Spatial dimension | Definition |
| --- | --- | --- | --- |
| $D_{short}$ | PET/CT | 1D | Diagnostic feature, maximum short diameter of the axial section |
| Area | PET/CT | 2D | Diagnostic feature, area of the axial section |
| Volume | PET/CT | 3D | Diagnostic feature, volume of the lymph node |
| $CT_{mean}$ | CT | 2D/3D | Diagnostic feature, mean CT value inside the lymph node |
| $CT_{contrast}$ | CT | 2D/3D | Diagnostic feature, the difference between $CT_{mean}$ and the mean CT value of a 2-mm-thick tissue layer surrounding the lymph node. |
| $SUV_{mean}$ | PET | 2D/3D | Diagnostic feature, mean SUV inside the lymph node |
| $SUV_{max}$ | PET | 2D/3D | Diagnostic feature, max SUV inside the lymph node |
| $SUV_{std}$ | PET | 2D/3D | Diagnostic feature, standard deviation of SUV inside the lymph node |
| 1st-order texture features | PET/CT | 3D | Six texture features calculated based on the pixel intensity histogram, see the supplementary material of [22] for detailed definition |
| 2nd-order texture features | PET/CT | 2D | Nineteen texture features calculated based on gray-level co-occurrence matrix (GLCM), see the supplementary material of [22] for detailed definition |
| High-order texture features | PET/CT | 3D | Five texture features calculated based on neighborhood gray-tone difference matrix (NGTDM) and 11 texture features calculated based on gray-level zone size matrix (GLZSM), see the supplementary material of [22] for detailed definition |

For the column of "spatial dimension", the term "2D/3D" means the feature is calculated for both 2D and 3D images for defining fuzzy boundaries. To reduce inter-rater variance, the final segmentation was proofread by a doctor with over 10 years experience. From the segmented slices, 3D volumes of lymph nodes were reconstructed.

The texture features were calculated following the instructions of a recent survey of PET/CT image characterization [22]. The texture features included 5 first-order features based on histogram analysis, 19 second-order features based on gray-level co-occurrence matrix (GLCM), and 16 high-order features based on neighborhood gray-tone difference matrix (NGTDM) and gray-level zone size matrix (GLZSM). The same texture features were calculated for both PET and CT, with 41 features for each modality. To avoid implementation variance, we used the publicly available code provided by the authors of [22]. Before texture calculation, the images were resampled to $1.0 \times 1.0 \times 1.0$ mm$^3$ isotropic voxel size. Different parameter values for calculating texture features were tested, the optimal values were selected by maximizing the AUC result. We used 64 bins for both PET and CT images, within CT HU range [−300, 1000] and PET SUV range [0, 20]. The pixel distances were 1 pixel, since most of the lymph nodes had a diameter shorter than 4 mm. The GLCM was calculated in 3D space, and the average matrix of all 13 neighborhood directions were used to calculate the features.

For comparison purpose, we subdivided the 95 features into four sets, i.e., D13 (13 diagnostic features), T82 (82 textural features), A95 (the combination of all 95 features), and S6 (6 selected features from A95). S6 was derived using the sequential forward feature selection method as in. The feature selection strategy was frequently used for classical machine learning methods like SVM to reduce the feature set and to improve classification accuracy. The four feature sets will be compared for each classical method.

### Validation strategy

For method validation, a nested cross-validation (CV) was performed. We used the outer CV loop to train and test the machine learning methods and used the inner CV loop to tune the method parameters. The outer loop contained 10 times 10-fold cross-validation, and the inner loop contained ninefold cross-validation using the training samples of the outer loop, a grid search was conducted to derive the optimal parameters for each method. To keep the balance between the positive and negative samples, 120 positive and 120 negative samples were randomly selected for each cross-validation. The CV was conducted on a per-node basis, so that different folds did not contain samples from the same lymph node. However, such strategy may assign different lymph nodes of the same patient to different folds. When this happened, the samples were exchanged between the folds so that all the samples of one patient only exist in one fold. In this way, we ensured the training and testing data do not contain samples from the same lymph node or the same patient, meanwhile keeping a balanced number of positive and negative samples in each fold. For fair comparison, different machine learning methods were trained with the same training sets and tested with the same testing sets.

For each cross-validation, we calculated the performance values for the five machine learning methods and the doctors. The performance values include sensitivity (SEN), specificity (SPC), diagnostic accuracy (ACC), and the area under the receiver operating characteristic (ROC) curve (AUC). The corresponding pathology analysis results were deemed as gold standard. The SEN, SPC, and ACC values were determined from the optimal cut point of the ROC curve, i.e., the point closest to (0,1). The doctors' performance was also evaluated using the same criterion. Because doctors only made binary diagnosis (positive or negative), no AUC could be calculated for human doctors.

Comparison between different feature sets and different methods was mainly performed based on the AUC and ACC values. Because the doctors do not have AUC results, the comparison between doctors and machine learning methods was only based on the ACC values. Due to the 10 times 10-fold CV, 100 groups of performance values were calculated for each feature set and each method; therefore, paired hypothesis tests of 100 samples were performed. The $p$ values were calculated using paired $t$ test if the samples were normally distributed; otherwise, the Wilcoxon-signed rank test was used. Multiple comparison correction strategies were applied to limit the accumulation of false positives of multiple tests. We applied two types of widely used correction strategies, i.e., the Bonferroni correction which controls false positives but potentially increases false negatives and the false discovery rate (FDR) correction (at $q = 0.05$) which generates less false negatives at the cost of increased false positives than the Bonferroni correction. The null hypotheses were rejected at the level of $p < 0.05$ after correction.

We firstly compared the four feature sets (D13, T82, A95, and S6) for each classical method and selected the optimal feature set with the highest mean AUC across 100 CVs. Afterwards, the classical methods with their optimal feature sets were compared with CNN and human doctors.

### Results

Table 3 reports the performance values of the machine learning methods and human doctors. For the four classical methods, the results of each feature set are also listed. All the performance values are listed as the means and standard deviations of the 100 cross-validations.

Table 3 Performance values of the machine learning methods and human doctors

|  | SEN/% | SPC/% | ACC/% | AUC |
|---|---|---|---|---|
| SVM+D13 | 77.08 ± 17.97 | 88.37 ± 11.43 | 82.73 ± 9.97 | 0.9148 ± 0.0856 |
| SVM+T82 | 88.05 ± 14.80 | 50.42 ± 14.89 | 69.23 ± 8.57 | 0.7997 ± 0.1106 |
| SVM+A95 | 77.08 ± 17.97 | 88.37 ± 11.43 | 82.73 ± 9.97 | 0.9148 ± 0.0856 |
| SVM+S6 | 77.82 ± 17.36 | 88.48 ± 10.77 | 83.15 ± 10.13 | 0.8997 ± 0.0927 |
| RF+D13 | 81.56 ± 16.71 | 88.59 ± 10.55 | 85.08 ± 9.88 | 0.9165 ± 0.0764 |
| RF+T82 | 73.35 ± 16.55 | 82.66 ± 11.17 | 78.01 ± 10.03 | 0.8624 ± 0.0855 |
| RF+A95 | 84.91 ± 14.52 | 84.06 ± 12.52 | 84.49 ± 9.16 | 0.9137 ± 0.0758 |
| RF+S6 | 82.89 ± 15.76 | 85.12 ± 10.92 | 84.00 ± 9.59 | 0.9031 ± 0.0867 |
| AdaBoost+D13 | 85.65 ± 13.46 | 84.45 ± 12.61 | 85.05 ± 8.97 | 0.9143 ± 0.0751 |
| AdaBoost+T82 | 70.73 ± 19.14 | 81.58 ± 11.81 | 76.15 ± 11.22 | 0.8466 ± 0.0903 |
| AdaBoost+A95 | 82.95 ± 14.45 | 84.56 ± 12.24 | 83.76 ± 9.36 | 0.9097 ± 0.0741 |
| AdaBoost+S6 | 84.08 ± 15.38 | 86.62 ± 10.51 | 85.35 ± 8.86 | 0.9044 ± 0.0802 |
| BP-ANN+D13 | 75.65 ± 19.61 | 85.38 ± 12.03 | 80.51 ± 11.01 | 0.8740 ± 0.0978 |
| BP-ANN+T82 | 53.26 ± 36.20 | 71.02 ± 31.77 | 62.14 ± 14.61 | 0.6581 ± 0.1831 |
| BP-ANN+A95 | 51.89 ± 38.60 | 72.87 ± 31.89 | 62.38 ± 15.27 | 0.6458 ± 0.2281 |
| BP-ANN+S6 | 75.62 ± 16.48 | 75.91 ± 16.80 | 75.76 ± 10.83 | 0.8117 ± 0.1115 |
| CNN | 83.53 ± 13.85 | 87.75 ± 10.30 | 85.64 ± 8.09 | 0.9086 ± 0.0865 |
| Doctor | 72.93 ± 13.10 | 90.35 ± 9.87 | 81.61 ± 7.86 | — |

For each classical method, the results of each feature set are listed. The row of best feature set is marked with gray background

### Feature comparison

The optimal feature set of each classical method was selected according to the AUC and ACC values. Figure 1 plots the AUC and ACC values of different feature sets of each method. As shown by Table 3 and Fig. 1, D13 is the optimal feature set of each classical method, while T82 is the worst one. For RF, SVM, and AdaBoost, T82 yields lower AUC and ACC than any other feature set

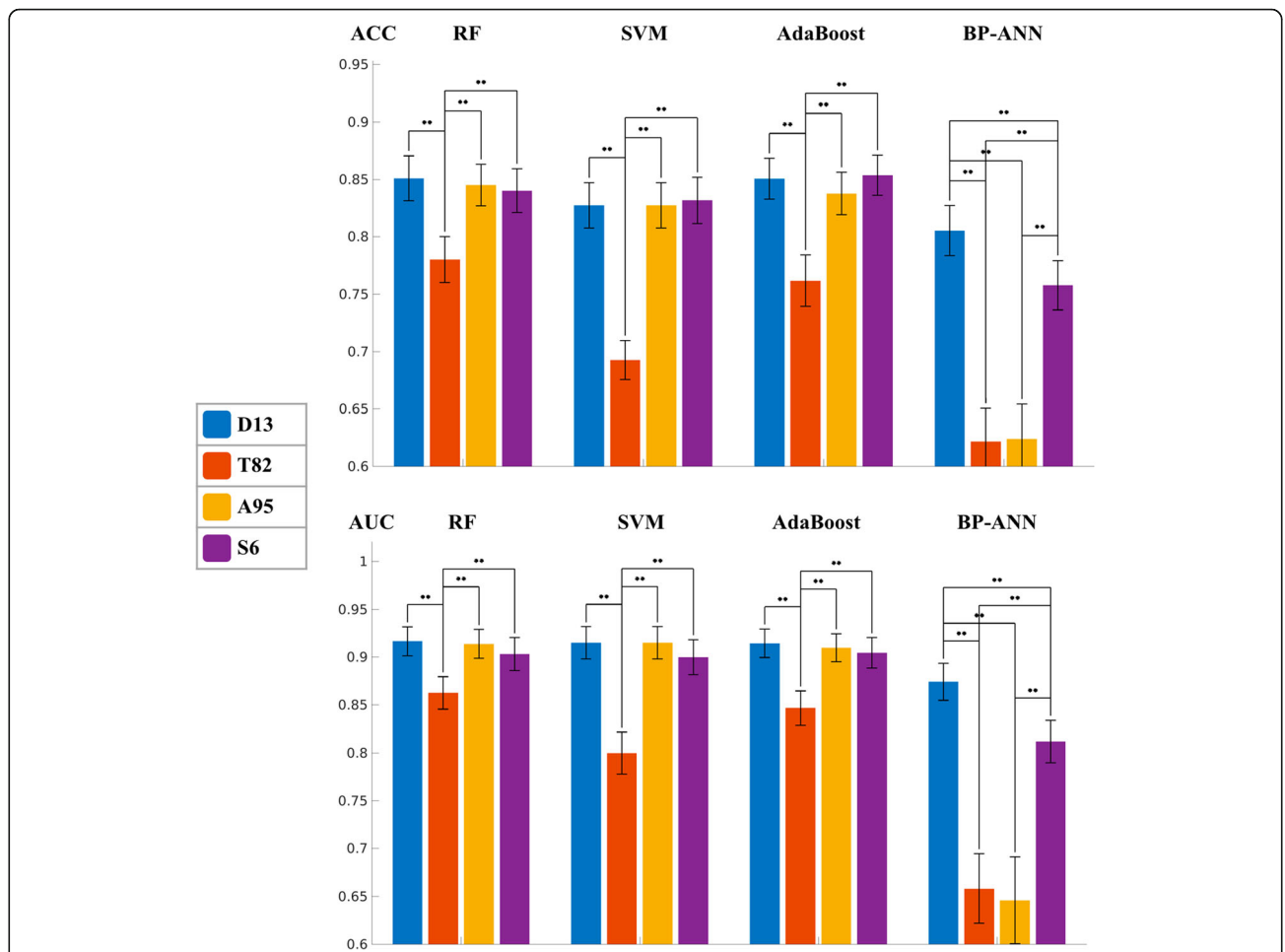

**Fig. 1** Comparison between different feature sets of the four classical machine learning methods, based on mean AUCs and mean ACCs of the 10 times 10-fold cross-validation. The error bars indicate 95% confidence interval. The *p* value between different feature sets are plotted as *bridge* and *stars*, where *two stars* means $p < 0.05$ after both Bonferroni and FDR corrections, and *one star* means $p < 0.05$ only after FDR correction

($p < 0.05$ after both Bonferroni and FDR corrections). For most cases, the differences are not evident between D13, A95, and S6, except for A95 vs. D6 for SVM ACC. However, for BP-ANN, the AUC and ACC of A95 is as low as T82, implying that adding T82 to D13 dramatically deteriorates the performance of BP-ANN. Comparing the four classical methods based on each feature set, AdaBoost, SVM, and RF are better than BP-ANN, while the difference between AdaBoost, SVM, and RF is not evident. These findings agree with a comprehensive study [36] of different classical machine learning methods for quantitative radiomic biomarkers.

**Method comparison**

Since D13 is the optimal feature set for all classical methods, we used the performance values of D13 to compare the classical methods with CNN and human doctors. Figure 2 displays the mean values and confidential intervals of AUC and ACC of each method. It could be observed that AdaBoost is the method of the highest AUC, and CNN is the method of the highest ACC. BP-ANN is the method of the lowest AUC and ACC, and it is worse than any other method in terms of AUC. The differences of AUC between BP-ANN and all other methods were significant ($p < 0.05$) after FDR correction. After the Bonferroni correction, the differences were only significant between BP-ANN and RF and between SVM and AdaBoost. SVM demonstrates no evident differences with RF, AdaBoost, and CNN in terms of AUC and ACC. Compared to ANN, SVM's AUC value is significantly higher ($p < 0.05$ after both Bonferroni and FDR corrections), while the ACC value is higher but not significant.

Because human doctors do not have AUC results, we only compared the doctors with machine learning methods based on ACC. Figure 2 indicates that CNN, RF, and AdaBoost have better ACC than the doctors, but the differences were not significant after the Bonferroni

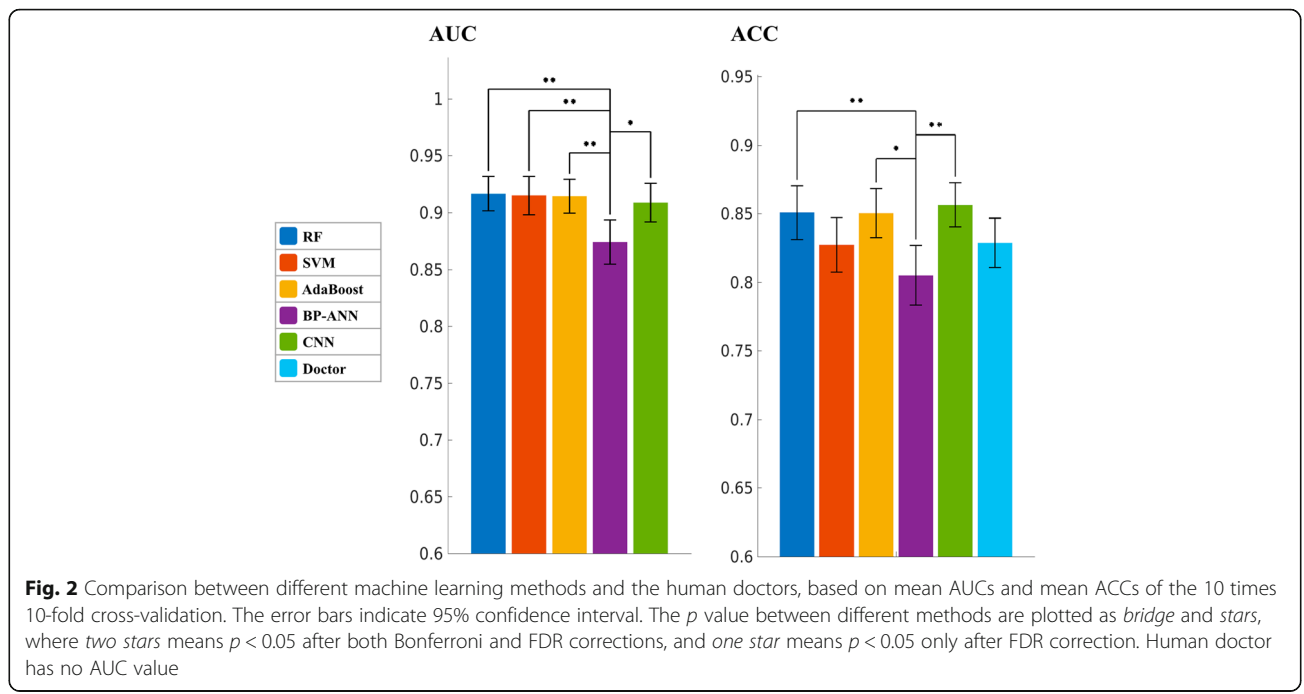

**Fig. 2** Comparison between different machine learning methods and the human doctors, based on mean AUCs and mean ACCs of the 10 times 10-fold cross-validation. The error bars indicate 95% confidence interval. The *p* value between different methods are plotted as *bridge* and *stars*, where *two stars* means $p < 0.05$ after both Bonferroni and FDR corrections, and *one star* means $p < 0.05$ only after FDR correction. Human doctor has no AUC value

and FDR corrections. The *p* value between CNN and human was less than 0.05 before correction; however, it did not survive the FDR or Bonferroni correction. To make further comparison, the ROC curves of all machine learning methods are plotted in Fig. 3, together with the performance point of human doctors. The ROC curve of each method is plotted as the average curve of 100 cross-validations. As revealed by Fig. 3, at the sensitivity level of human doctors, CNN, RF, and AdaBoost have better specificity than the doctors, and SVM has close specificity to the doctors. However, when higher sensitivity or higher specificity is considered, SVM quickly catches up with CNN, RF, and AdaBoost. In contrast, the curve of BP-ANN is always below others.

Comparing the timing performance of the five methods, SVM is the fastest, it took ~3 s to train for each fold based on a A95 feature set, using a computer with 2.2 GHz dual core i7 CPU. For the same training data, RF and AdaBoost took ~6 and ~40 s, respectively. BP-ANN took ~1 h to train using a CPU or ~2 min to train using a GPU acceleration based on NVIDIA Quadro K1100M graphics card. CNN was trained with the GPU acceleration on a workstation with 449 G RAM and NVIDIA Tesla K40C graphics card; it took ~10 min for each fold of training. For testing, all the methods took less than 1 s for a single cross-validation.

## Discussion

### Feature comparison

In recent years, an increasing number of publications were using PET/CT texture features for tumor classification. On the other side, some studies claim that texture features are not reliable because their values are influenced by factors irrelative to the tumor's genuine property, such as tumor size and image processing procedures [21]. In this study, we made efforts to follow the suggested workflow of texture feature calculation by a recent review study [22] and used their published code to avoid implementation bias. However, texture features still performed significantly worse than the diagnostic features. The main reason for the unsatisfactory performance of texture features might be the small sizes of the lymph nodes. According to Table 1, 80.3% of the tested lymph nodes had a short axis diameter less than 4 mm. Small tumor size means insufficient number of voxels for meaningful heterogeneity measurement; therefore, the discriminative power of the texture features was compromised. Because such small-sized lymph nodes dominate in a clinical scenario, texture features might not be suitable for classifying a small size mediastinal lymph node from PET/CT images.

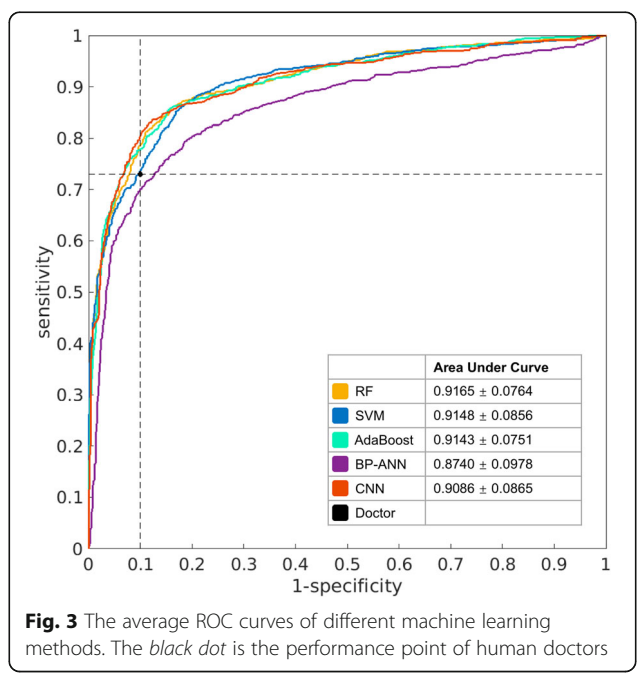

**Fig. 3** The average ROC curves of different machine learning methods. The *black dot* is the performance point of human doctors

Method comparison

As shown by the comparison of classical results, AdaBoost, RF, and SVM performed better than ANN in terms of both AUC and ACC. From a methodology aspect, AdaBoost and RF are both ensembles of decision trees. We found that some other comparison studies [36, 37] also demonstrated that ensemble methods outperform other classifiers. The mechanism of a decision tree can utilize different features to compensate each other, and the ensemble of decision trees combines multiple weak tree classifiers into a strong classifier. As a result, the ensembles of decision trees can yield good classification results based on an even weak feature set. As shown by Table 3 and Fig. 1, when only T82 was used as input features, AdaBoost and RF yielded better AUC and ACC than SVM and BP-ANN. SVM belongs to the kernel-based classifier family, which implicitly maps the input features into a higher dimensional feature space using a kernel function that measures the distance between feature points in the mapped space. By such kernel-based mapping, SVM is able to achieve much better classification performance than conventional linear classification methods. The choice of kernel function greatly affects the performance of SVM, and the nonlinear kernel function used in this study helped SVM to maintain good performance even with suboptimal input features (A95 and S6).

ANN and CNN both belong to the neural networks family, but ANN performed worse than CNN. We only used two hidden layers for ANN; it seems that the imperfect performance of ANN might be due to the insufficient number of layers. However, as we tested different layer numbers (from one hidden layer to seven hidden layers), the best number was two instead of seven. Although it is generally assumed that deeper networks perform better than shallower ones, such assumption is valid when there is enough training data, while the training method should also be good at learning deep networks. In this study, the training data of BP-ANN is not abundant enough to support a deeper ANN, and the back-propagation method is not suitable for training deep networks [33]. In contrast, CNN is well designed for learning deep networks, and it also uses data augmentation to increase the training set.

Compared with human doctors from our institute, all the five machine learning methods had higher sensitivity but lower specificity than human doctors. Doctors tended to underestimate the malignant tumors because most of the lymph nodes in this study were small in size. The machine learning methods gained better sensitivities than human doctors at the cost of losing specificities. When ACC was used as more balanced criteria than sensitivity and specificity, RF, AdaBoost, and CNN were better than human doctors, but the difference was not significant after Bonferroni and FDR corrections.

In many recent publications of medical image analysis, CNN was reported to outperform classical methods for imaging modalities other than PET/CT. In this study, CNN is not significantly better than RF, AdaBoost, or SVM, because it has not fully explored the functional nature of PET. Before the image patches are input to CNN, the pixel intensities are normalized to a range of [−1, 1], thus the discriminative power of SUV is lost during the normalization. It was surprising to find that without the important SUV feature, the difference between CNN and the best classical methods is not evident. CNN utilizes the image appearance pattern around the lymph node. The appearance pattern includes information of local contrast, nearby tissues, boundary sharpness, and etc. Such information is different from but as powerful as the diagnostic features like SUV, tumor size, and local heterogeneity. To illustrate the discriminative power of the image appearance pattern, we extracted the intermediate feature vector produced by the internal flattening layer of the CNN. This was a vector of 512 features, which could be considered as a sparse representation of the image patch's appearance. We used the 512 features as the input of the classical methods. For RF, the 512 features resulted in AUC and ACC of 0.89 and 80.8%, respectively. For SVM, the AUC and ACC were 0.89 and 80.6%, respectively. These results were much higher than the AUC and ACC of T82, and they were even close to the results of D13. Unlike the texture features, the appearance patterns of CNN are not affected by the size of the lymph nodes, because they are computed from the entire image patch which includes both the lymph node and its surrounding tissues. Therefore, the image appearance pattern can be a promising substitute for the texture features, as well as a good compensation to the diagnostic features.

This study used the AlexNet for CNN architecture, but with a reduced number of layers. The reason for using less neuron layers was to avoid overfitting to the training data. Although we used 729 times data augmentation, the total number of training data for each cross-validation was still not abundant compared to many other deep learning applications. This was also the reason that we did not use the more advanced CNN architectures like VGGNet [38], GoogleNet [39], and ResNet [40] which were designed for a much larger training set. In future work, if we can collect more data from multi-centers, deeper CNN architectures will be explored. Recently, there are some studies using a small set of medical images to fine-tune the deep network learned from a large natural image set, in order to solve the problem of insufficient medical training data [41]. However, it is to be investigated if this method could perform well on PET images, since the appearance of PET is quite different from natural images.

In this study, image patches of both modalities (PET and CT) were mixed into the same network. Such mixed setting may potentially limit the performance of CNN, because the PET and CT patches contained different types of diagnostic information. It should be more appropriate to process the PET and CT patches with separated subnetworks and combine the results of different subnetworks at the output layers. However, there is currently no such architecture for dual-modality PET/CT images, we will leave this issue for future research.

## Conclusions

In conclusion, this study revealed that the diagnostic features are more discriminative than the texture features, mainly because the calculation of texture features is not reliable due to the small lymph node size. CNN is a recently popular method which utilizes image appearance patterns for classification. In this study, the performance of CNN is not significantly different from the best classical methods, even though it did not use the important diagnostic features like SUV and tumor size. Moreover, CNN does not take hand-crafted features as input, it eliminated the needs for tumor segmentation and feature selection, making the whole process much more convenient and less prone to user bias. CNN also avoids using the debated texture features which are affected by tumor size. Our future direction will focus on improving the CNN performance by incorporating diagnostic features into the network, as well as designing more dedicated network structure for dual-modality PET/CT images. This study was a single-center retrospective study. For the future, we are planning to collect multi-center data to conduct more generalize evaluation, as well as to explore the potential of deep learning with more training data.

**Acknowledgements**
This research was supported by the general program of National Natural Science Fund of China (Grant No. 61571076),the youth program of National Natural Science Fund of China (Grant No. 81401475), the general program of National Natural Science Fund of China (Grant No. 81171405), the general program of National Natural Science Fund of China (Grant No. 81671771), the general program of Liaoning Science & Technology Project (Grant No. 2015020040), the cultivating program of Major National Natural Science Fund of China (Grant No. 91546123), and the Basic Research Funding of Dalian University of Technology (Grant No. DUT14RC(3)066). The authors thank Dr. Larry Pierce for sharing the publicly available codes for computing PET/CT texture features.

**Authors' contributions**
LY, WL, and HW together developed the study concept. HW programmed the methods for cross-validation and diagnostic feature calculation, and wrote the manuscript. WL prepared the data samples. ZZ conducted research on CNN, calculated the texture features and collected the validation results of all tested methods. ZC fine-tuned parameters of classical ANN method. YL, PL, WW, and LY collected the patient images, made the doctor diagnosis, conducted the pathology analysis, specified the lymph node center, and performed image segmentation. LY also critically reviewed the manuscript. All authors approved the final manuscript.

**Competing interests**
The authors declare that they have no competing interests.

**Ethics approval and consent to participate**
All procedures performed in studies involving human participants were in accordance with the ethical standards of the institutional and/or national research committee and with the 1964 Helsinki declaration and its later amendments or comparable ethical standards. For this type of study, formal consent is not required.

**Author details**
[1]Department of Biomedical Engineering, Faculty of Electronic Information and Electrical Engineering, Dalian University of Technology, No. 2 Linggong Street, Ganjingzi District, Dalian, Liaoning 116024, China. [2]Department of Biomedical Informatics and the College of Health Solutions, Arizona State University, 13212 East Shea Boulevard, Scottsdale, AZ 85259, USA. [3]Center of PET/CT, The Affiliated Tumor Hospital of Harbin Medical University, 150 Haping Road, Nangang District, Harbin, Heilongjiang Province 150081, China. [4]HIT–INSA Sino French Research Centre for Biomedical Imaging, Harbin Institute of Technology, Harbin, Heilongjiang 150001, China.